\documentclass{article}
\usepackage{spconf}
\usepackage[T1]{fontenc}
\usepackage{amsmath,amssymb,graphicx,booktabs,url}
\newcommand{\ind}{\mathbf{1}}
\title{FUNCTION-STRUCTURED REINFORCEMENT LEARNING WITH EXECUTABLE VERIFIERS FOR MATHEMATICAL REASONING}
\name{
  Zihan Liu$^{1}$ \qquad Xurong Xie$^{2,*}$
  \thanks{
    \fontsize{9pt}{10.5pt}\selectfont
    $^{*}$Corresponding author: Xurong Xie.
  }
}
\address{
$^{1}$Department of Computer Science, University College London, UK\\
$^{2}$Institute of Software, Chinese Academy of Sciences, China\\
$^{1}$\texttt{ucabz62@ucl.ac.uk}
\qquad
$^{2}$\texttt{xurong@iscas.ac.cn}\\
}

\begin{document}
\ninept
\raggedbottom
\maketitle
\setlength{\abovedisplayskip}{6pt plus 2pt minus 2pt}
\setlength{\belowdisplayskip}{6pt plus 2pt minus 2pt}
\setlength{\abovedisplayshortskip}{0pt plus 2pt}
\setlength{\belowdisplayshortskip}{4pt plus 2pt minus 2pt}
\begin{abstract}
Algorithmic mathematical reasoning requires reliable decomposition,
computation, and aggregation. Final-answer rewards provide limited
guidance on intermediate errors, while successful execution does
not guarantee mathematical correctness.
This work proposes Function-Structured Graph Reinforcement
Learning (FSG-RL), connecting subproblem graphs and Python
implementations with multi-verifier feedback.
The policy first learns to generate code from function graphs
through supervised fine-tuning (SFT).
Group Relative Policy Optimization (GRPO) then optimizes the
policy using answer-gated rewards and span-level credit assignment.
The framework also supports teacher supervision and structured memory.
A benchmark curated from Grade School Math 8K (GSM8K), MathQA,
MATH, and Omni-MATH pairs public function graphs with private
verification specifications.
Under a unified evaluation protocol, GRPO improves final-answer
accuracy from 43.25\% to 67.50\% and full solution success from
32.25\% to 52.25\% over SFT.
Continued reinforcement learning (RL) with teacher supervision
yields additional gains.
The gains extend beyond producing correctly formatted code,
supporting verifier-guided reinforcement learning for
mathematical reasoning.
Code is available at
\url{https://github.com/ZihanLiummyycc/FSG-RL}.
\end{abstract}

\section{Introduction}

This study examines mathematical problems that require reasoning and
computation. Problems are selected from four datasets to cover
different types and difficulty levels: multistep arithmetic
in GSM8K~\cite{cobbe2021gsm8k}, word problems involving mathematical operations in MathQA~\cite{amini2019mathqa}, algebra, number theory,
counting, and probability in selected Level 1--3 MATH
problems~\cite{hendrycks2021math}, and more demanding
competition problems in Omni-MATH~\cite{gao2024omnimath}.
Solving these problems requires identifying useful subproblems,
implementing their computations, and combining intermediate
results into a final answer. 

\begin{figure*}[t]
\centering
\includegraphics[width=0.99\textwidth]{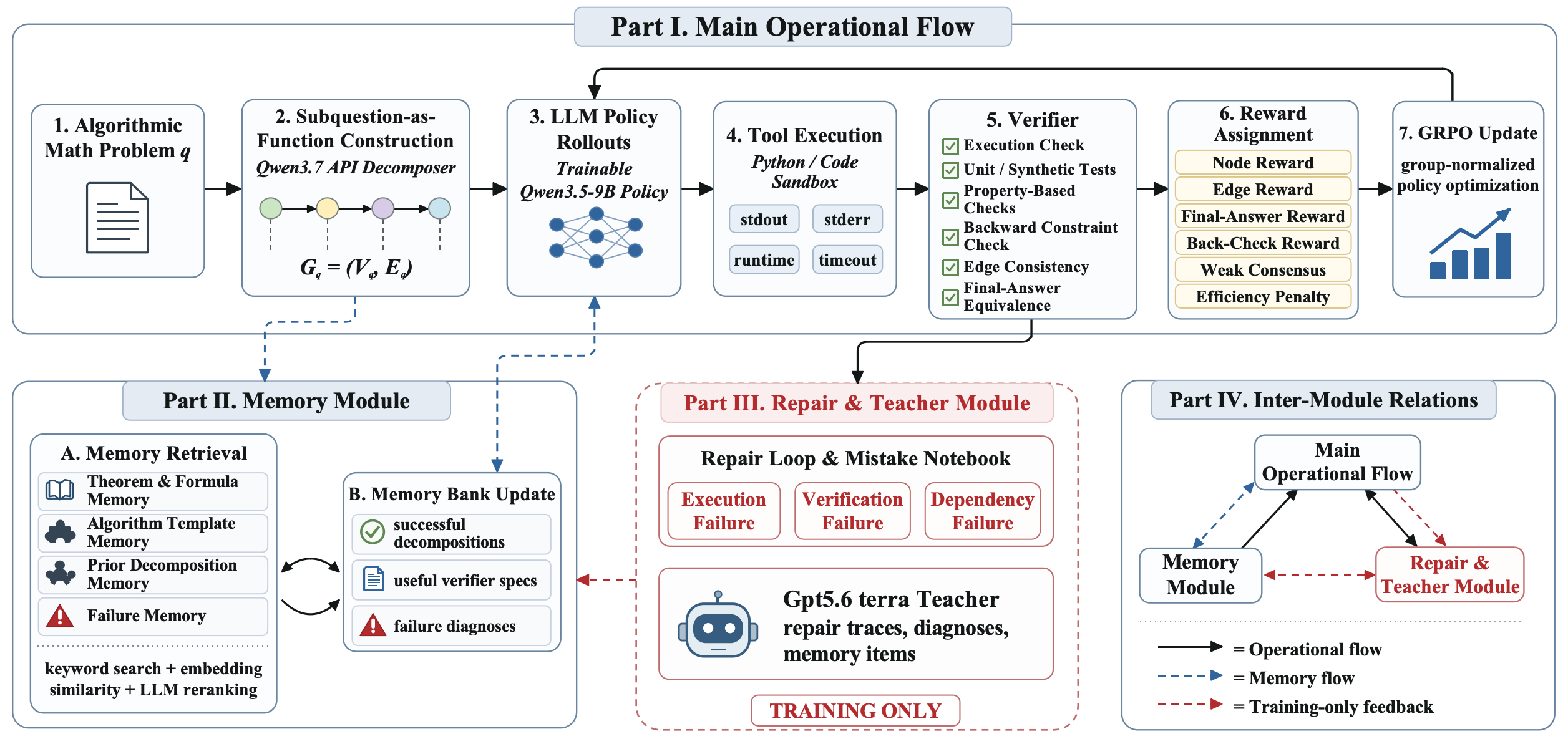}
\caption{Overview of FSG-RL. The core pipeline connects function
graphs, executable solutions, and verifier-guided learning.
Memory retrieval and teacher supervision provide modular extensions.}
\label{fig:framework}
\end{figure*}

Prior work has improved how models organize complex reasoning.
Least-to-most prompting~\cite{zhou2023leasttomost} breaks a problem
into simpler subproblems and solves them sequentially using earlier
answers. Graph of Thoughts~\cite{besta2024graphofthoughts} extends
this structure to graphs that support aggregation and refinement.
These approaches make reasoning dependencies explicit but lack testable function interfaces for intermediate computations. DeepSeek-Prover-V2~\cite{ren2025deepseekproverv2}
connects subgoal decomposition with formal verification in Lean.
It demonstrates the value of checkable subgoals through formal proof verification, whereas this work checks the behavior of Python implementations.

Tool-integrated methods address the computational side of reasoning. PAL (Program-aided Language Models)~\cite{gao2023pal} delegates computation to a Python interpreter,
while ReAct (Reasoning and Acting)~\cite{yao2023react} interleaves reasoning and actions
to incorporate environmental feedback. ToRA (Tool-integrated Reasoning Agents)~\cite{gou2024tora} develops tool-integrated mathematical
reasoning, and MathCoder~\cite{wang2024mathcoder} trains models
on interleaved natural language, code, and execution results.
ReTool~\cite{feng2025retool} further uses outcome based reinforcement
learning to learn when and how to invoke computational tools.
These methods reduce the burden of computation and improve tool
use. However, an execution result alone does not establish whether a function correctly implements the intended computation or whether its output is correctly combined with other results. This motivates checks
of intermediate behavior and subproblem consistency.

A complementary line of work improves training feedback.
DeepSeekMath~\cite{shao2024deepseekmath} introduces GRPO for reinforcement learning in mathematical reasoning. Lightman et al.~\cite{lightman2024verify} show the value of
supervising intermediate reasoning steps through learned reward
models. Subproblem Curriculum Reinforcement Learning (SCRL)~\cite{jiang2026scrl} constructs verifiable subproblems
and assigns normalized advantages to their corresponding answer
spans. These advances provide finer learning signals than a final
outcome alone. For executable mathematical reasoning, the remaining
question is how to connect such local credit to function behavior,
dependency consistency, and final-answer correctness.

This paper proposes Function-Structured Graph Reinforcement
Learning (FSG-RL), a framework that connects these
elements. Subproblems form a typed dependency graph, with
public Python interfaces for executable nodes and private
verification specifications. Under this structure, the policy is first trained through SFT to construct function graphs and generate code, and then optimized with GRPO using rewards derived from program execution, unit and property tests, dependency checks, and final outcome verification. Answer gating reduces the weight of intermediate step rewards when the final answer is incorrect. Furthermore, Span-level credit assignment provides verification based learning signals for the corresponding code spans.
Our contributions are threefold:
\begin{itemize}
\setlength{\itemsep}{0pt}
\setlength{\parsep}{0pt}
    \item This paper proposes a novel function-structured framework
    for mathematical reasoning. It connects subproblem graphs,
    Python implementations, and multi-verifier feedback, integrating
    staged SFT-to-RL training with modular teacher supervision
    and structured memory.

    \item A curated benchmark for executable mathematical reasoning is released. Public function graphs and private verification specifications support an evaluation of interface compliance, functional correctness, and complete solution success.

    \item The experiments show that verifier guided GRPO improves solution
    correctness beyond the interface compliance established by SFT.
    The trained Qwen3.5-9B policy outperforms the Qwen3.8-27B application programming
    interface (API) reference for all
    outcome metrics. Teacher supervision
    provides additional observed gains.
\end{itemize}

\section{Method}
\subsection{Function Graphs and Private Verification}
\label{sec:function_graphs}

Function graphs connect subproblems, executable implementations,
and private verification. Part~I of Fig.~\ref{fig:framework}
shows the core pipeline from graph construction to
verifier-guided GRPO.
Annotation, teacher, and execution settings are specified
in Secs.~\ref{sec:framework_eval}--\ref{sec:extensions_eval}.

For a problem $q$ with gold answer $a^\star$, an annotation
pipeline constructs an acyclic graph $G=(V,E)$.%
Each public node specifies an identifier, a subquestion,
an output type, and a Python signature when executable.
Every node must contribute to the main answer. Edges describe
dependencies such as value use, formula implementation,
and aggregation. The policy receives $x=(q,G)$ and produces
\begin{equation}
y=\bigoplus_{v\in V}
\texttt{<a\_v>}\;y_v\;\texttt{</a\_v>},
\label{eq:format}
\end{equation}
where $\bigoplus$ denotes concatenation in the prescribed
node order. Each executable span contains one self-contained
Python block, and the main span ends with a boxed answer.

Private verification checks the response after generation.
Unit and property tests check function behavior, while the output of a specified function call is compared with the reported answer. During dataset construction, we verify that reference implementations pass their tests, faulty variants fail, and reference target calls return answers equivalent to the gold answers. These checks provide a limited quality filter. The optional memory module in Part~II of
Fig.~\ref{fig:framework} retrieves Mathlib4 theorem
statements~\cite{mathlib2020} enriched with natural-language descriptions.
It supports retrieval augmented generation~\cite{lewis2020rag}
by combining keyword matching with embedding cosine
similarity~\cite{reimers2019sbert} to retrieve the top three
entries from each memory category.

\subsection{Multi-Verifier Rewards}

The reward evaluates three aspects: interface validity, local
functional correctness, and final outcome correctness.
Let $M\in\{0,1\}$ indicate valid output tags, and let
$S,E,U,P,D\in[0,1]$ denote signature compliance, execution success,
unit-test pass rate, property-test pass rate, and dependency evidence,
respectively.
Property tests check broader behavioral
invariants~\cite{claessen2000quickcheck}. Let $\mathcal E$ denote the graph's edge set. Dependency evidence is
\begin{equation}
D=\frac{1}{|\mathcal E|}\sum_{e\in\mathcal E}d_e,
\label{eq:dependency}
\end{equation}
with $D=0$ for an empty edge set.
Each $d_e$ uses a relation specific check. Value use edges measure whether extracted source values appear in the target text or outputs.
Formula edges use the target's property test pass rate, falling back
to its overall test pass rate. Aggregation edges check whether a
source test output matches the target's reported answer.
Output equivalence edges compare the two output records.
These checks provide auxiliary consistency evidence.
Scores $E,U,P$ are averaged over applicable nodes, so nodes with
more tests do not receive greater global weight.

Answer correctness differs from program answer consistency.
For the extracted answer $\hat a$, gold answer $a^\star$, and
program output $c$ on the private target call,
\begin{equation}
Y=\ind[\hat a\equiv a^\star],\qquad
B=\ind[c\equiv\hat a].
\label{eq:outcome}
\end{equation}
The program output may match the reported answer even when both are incorrect. Thus, $B$ alone cannot establish correctness and is used alongside $Y$.
The weights encode preset priorities.
Format and signature checks receive small weights because they
enable verification without establishing correctness.
Execution and property tests provide complementary local evidence.
Unit tests receive greater weight for checking explicit input with output
contracts, while dependency evidence receives a small auxiliary weight.
Final answer correctness has the largest coefficient, supported
by a separate program answer consistency reward.

The answer gate is $g(Y)=\eta+(1-\eta)Y$, where
$\eta\in[0,1]$ controls the process credit retained for
incorrect answers. With nonnegative weights $\lambda_j$,
the total reward is
\begin{align}
R=M\big[&\lambda_M+\lambda_S S
+g(Y)(\lambda_E E+\lambda_U U \nonumber\\
&+\lambda_P P+\lambda_D D)
+\lambda_Y Y+\lambda_B B\big].
\label{eq:reward}
\end{align}
The experiments use $\eta=0.25$,
$\lambda_M=0.10$, $\lambda_S=0.15$,
$\lambda_E=\lambda_P=0.50$,
$\lambda_U=\lambda_B=0.75$,
$\lambda_D=0.05$, and $\lambda_Y=1.50$.
The gate preserves full process credit for correct answers
and one quarter for incorrect answers, retaining local
learning signals even in an all-wrong rollout group.

Under these settings, with component scores in $[0,1]$ and
no additional teacher reward, the core reward satisfies
\begin{align}
R_{\mathrm{wrong}}
&\le 0.25+0.25(1.80)+0.75=1.45,
\nonumber\\
R_{\mathrm{correct}}
&\ge 0.10+1.50=1.60,
\label{eq:reward_ordering}
\end{align}
where the lower bound requires valid output tags.
Thus, the chosen weights ensure that a format-valid correct
answer outranks any incorrect answer.
This is a total-reward ordering property, not a guarantee
about token advantages after span-level normalization.

\subsection{Span-Level Group Relative Optimization}
\label{sec:span_grpo}

The total reward evaluates a complete response. To assign local
credit, each node's verification scores are combined with adjacent
edge evidence and shared outcome credit.
For rollout $i$, let $Q_i=1.50Y_i+0.75B_i$. The reward for node $v$ is
\begin{align}
R_{iv}=M_i\Big[&
0.15S_{iv}+c_vQ_i \nonumber\\
&+g(Y_i)\Big(
0.50E_{iv}+0.75U_{iv}+0.50P_{iv} \nonumber\\
&\qquad+0.025\sum_{e\in\mathcal I(v)}d_{ie}
\Big)\Big],
\label{eq:span_reward}
\end{align}
where $\mathcal I(v)$ contains the edges incident to $v$.
Here, $c_v=1$ for the main node and $c_v=0.5$ otherwise.
Each edge contributes half its reward to each endpoint.
Unavailable components contribute zero. Local property scores may
also include declared formula or value checks.
Outcome credit is shared, so the span rewards need not sum to
the total reward.

For the $K$ rollouts of one problem, total rewards and each
node's rewards are normalized separately:
\begin{equation}
\mathcal N(z_i)=
\begin{cases}
0, & \sigma_z<10^{-8},\\
(z_i-\bar z)/\sigma_z, & \text{otherwise},
\end{cases}
\label{eq:advantage}
\end{equation}
where $\bar z$ and $\sigma_z$ are the group mean and population
standard deviation. Token advantages are then assigned as
\begin{equation}
A_{it}=
\begin{cases}
\mathcal N(R_{iv}), & t\in\mathcal T_{iv},\\
\mathcal N(R_i), & \text{otherwise},
\end{cases}
\label{eq:token_advantage}
\end{equation}
where $\mathcal T_{iv}$ contains the content tokens between the
opening and closing tags of node $v$.
Tags and other completion tokens retain the total reward advantage.
Prompt and padding tokens are excluded from the loss.

For token history $h_{it}=(x,y_{i,<t})$, define
\begin{equation}
\rho_{it}=
\frac{\pi_\theta(y_{it}\mid h_{it})}
{\pi_{\rm old}(y_{it}\mid h_{it})},
\qquad
\delta_{it}=
\log\frac{\pi_{\rm ref}(y_{it}\mid h_{it})}
{\pi_\theta(y_{it}\mid h_{it})}.
\end{equation}
Following the clipped PPO objective~\cite{schulman2017ppo},
each rank minimizes
\begin{align}
\mathcal L=\frac{1}{T}\sum_{i,t}\Big[
&-\min\big\{\rho_{it}A_{it},
\operatorname{clip}(\rho_{it},1-\epsilon,1+\epsilon)A_{it}\big\}
\nonumber\\
&+\beta_{\rm KL}
\big(e^{\delta_{it}}-\delta_{it}-1\big)
\Big],
\label{eq:loss}
\end{align}
where $T$ is the number of generated tokens in the local group.
The old policy is fixed before each group update, and the reference
policy remains frozen at the stage's initial checkpoint.
Each rank processes one problem with its complete $K$-rollout group.
Advantages are normalized within that problem, not across ranks.
Gradients from the token-normalized local losses are then averaged.

\section{Experiments and Analysis}

Our experiments examine how executable interfaces, verifier guided
training, and teacher supervision contribute to reliable mathematical
problem solving. The analysis first defines what constitutes a complete solution,
then compares training stages and teacher continuations.
An external API reference places the trained policies in context.

\subsection{Benchmark and Evaluation Protocol}
\label{sec:benchmark}
\label{sec:framework_eval}

The function-structured benchmark is curated from
GSM8K~\cite{cobbe2021gsm8k}, MathQA~\cite{amini2019mathqa},
selected MATH problems~\cite{hendrycks2021math},
and Omni-MATH~\cite{gao2024omnimath}.
Each problem is paired with a public function graph and
private verification specifications. Annotated graphs and
implementations support SFT, while executable checks provide
GRPO rewards.

The benchmark contains 866 GRPO training problems and
400 evaluation problems from the four sources.
Teacher continuations use a subset of the training problems.

Runnable code can still compute or aggregate results incorrectly.
Motivated by intermediate supervision~\cite{lightman2024verify},
the evaluation distinguishes execution, local correctness,
and complete solution correctness.
All models receive common public graphs and are assessed
using identical private checks.

The reported metrics are execution success, passing all hidden
tests, program--answer consistency, final-answer accuracy,
and full success.
Back.\ denotes $B$, agreement between the private target call
output and the reported answer.
Ans.\ denotes $Y$, equivalence between the reported and gold
answers. Agreement alone does not imply correctness.

Full success requires valid tags, complete code blocks,
matching signatures, successful execution, all hidden tests
passing, and $B=Y=1$.
The dependency score $D$ contributes to training rewards
but is excluded from this criterion.

Local policies use greedy decoding with a 2,048-token limit.
After generation, Python execution is limited to five seconds
per~node.

\subsection{Staged SFT to RL Training}
\label{sec:training_eval}

With executable output distinguished from complete solutions,
the next experiment asks whether verification signals can guide learning.
SFT teaches the policy to express solutions through the required
interfaces. GRPO then uses differences in implementation correctness to guide policy updates. Comparing the base, SFT, and GRPO
policies reveals how solution quality changes across these stages.

Qwen3.5-9B-Base undergoes graph construction and solving SFT on
798 Omni problems, followed by executable solving SFT with
158 code problems and 632 mixed records.
GRPO then uses the training set in Sec.~\ref{sec:benchmark}.
Low-rank adapters~\cite{hu2022lora} are trained with
4-bit QLoRA~\cite{dettmers2023qlora} on two A100 GPUs.
All RL stages sample four rollouts per problem at temperature
0.8 and top-$p$ 0.95.
Main GRPO uses a learning rate of $10^{-6}$,
$\epsilon=0.20$, and $\beta_{\mathrm{KL}}=0.05$. Each stage freezes its initial adapter as the KL reference.

\begin{table}[t]
\centering
\caption{Training-stage comparison on the evaluation set (\%).}
\label{tab:main}
\setlength{\tabcolsep}{2.4pt}
\begin{tabular}{lrrrrr}
\toprule
Model & Exec. & Tests & Back. & Ans. & Full \\
\midrule
Qwen Base
& 23.75 & 20.00 & 1.25 & 2.50 & 1.00 \\
Executable SFT
& 86.00 & 61.75 & 41.50 & 43.25 & 32.25 \\
FSG-RL
& 90.25 & 65.75 & 63.75 & 67.50 & 52.25 \\
\bottomrule
\end{tabular}
\end{table}

Table~\ref{tab:main} shows two distinct stages of improvement.
SFT largely resolves the base model's difficulty with the
structured representation. The base model's score is constrained by its difficulty following
the required interfaces and does not isolate mathematical
problem-solving ability.
Subsequent GRPO yields larger gains in final-answer accuracy and full success than in execution success, while format compliance changes little. The main gain extends beyond producing valid code.

The reward design provides a reasonable explanation.
Once most candidates follow the required format, format rewards
offer little discrimination within a rollout group.
Unit and property tests assess local implementations, whereas
$Y$ and $B$ assess the correctness and consistency of the reported answer. Answer gating limits the value of local success when the final answer is wrong. Retained process credit is intended to distinguish
partial progress within an all wrong group.
Span level credit directs these signals to the relevant
implementation segments, connecting subproblem level
learning~\cite{jiang2026scrl} to executable behavior.

An exact McNemar test of final-answer correctness favors
GRPO over SFT, with $p=1.72\times10^{-18}$.

\subsection{Teacher Supervision and Qwen API Comparison}
\label{sec:extensions_eval}

With the main policy established, the next experiment asks whether teacher
feedback can further improve independent solving.
Verifier rewards indicate which candidates perform better, while
a teacher diagnosis can further explain what to revise.

Following the teacher-supervision path in Part~III of
Fig.~\ref{fig:framework}, the main checkpoint is further
trained on 60 hard problems using feedback-conditioned
GRPO (FC-GRPO). If initial verification
fails, \texttt{gpt-5.6-terra} diagnoses the student attempt and
suggests revisions, with at most one logical teacher request per~problem. 

\begin{table}[t]
\centering
\caption{Teacher continuation and Qwen API comparison (\%).}
\label{tab:teacher}
\setlength{\tabcolsep}{2.4pt}
\begin{tabular}{lrrrrr}
\toprule
Model & Exec. & Tests & Back. & Ans. & Full \\
\midrule
FSG-RL
& 90.25 & 65.75 & 63.75 & 67.50 & 52.25 \\
FC-GRPO
& 90.75 & 67.00 & 63.75 & 69.00 & 53.50 \\
\midrule
Qwen3.8-27B API
& 54.75 & 52.00 & 52.50 & 52.50 & 49.75 \\
\bottomrule
\end{tabular}
\end{table}

Teacher feedback can guide exploration toward useful corrections,
while verification rewards effective revisions. Even with
continuation training on only 60 problems, FC-GRPO shows additional
gains under teacher free evaluation, as reported in
Table~\ref{tab:teacher}. 

Finally, these independently evaluated policies are compared
with an external API reference.
Qwen3.8-27B is evaluated through the Inferera API, with
one response requested per problem. The temperature is set to zero,
the generation limit to 2,048 tokens, and the request timeout
to 180 seconds.
Under these settings, both GRPO trained policies exceed the Qwen3.8-27b API reference in final-answer accuracy and full success, whereas the SFT policy remains below it. Together with the local stage comparison, this pattern highlights
the value of training beyond format compliance.
The gains extend from reporting correct answers to producing
implementations that pass verification and support those answers.

\subsection{Illustrative Case Study}
\label{sec:case_study}

Execution and answer consistency alone do not ensure correctness.
Figure~\ref{fig:illustrative_case} illustrates this through two
constructed solutions. Both programs execute successfully and produce outputs consistent
with their reported answers, but only the parity-aware solution
is correct.
An even branch test detects the missing condition.

\begin{figure}[t]
\centering
\setlength{\fboxsep}{3pt}
\fbox{%
\begin{minipage}{\dimexpr\columnwidth-2\fboxsep-2\fboxrule\relax}
\raggedright

\textbf{Problem.}
For $n$ uniform on $\{1,\ldots,2023!\}$, find
$\Pr[\gcd(n^n+50,n+1)=1]$.

\smallskip
\textbf{Shared graph.}
$\texttt{reduce}
\rightarrow\{\texttt{even},\texttt{odd}\}
\rightarrow\texttt{prob}
\rightarrow\texttt{main}$\par

\smallskip
\textbf{Reduction.}
Since $n\equiv-1\pmod{n+1}$,
\[
\gcd(n^n+50,n+1)=\gcd(n+1,50+(-1)^n).
\]
The condition has period 714, which divides $2023!$.
The faulty path instead uses modulus 49 for both branches.

\smallskip
{\centering
\setlength{\tabcolsep}{3pt}
\begin{tabular}{lcc}
\toprule
& Parity-aware & Parity-omitting \\
\midrule
Even/odd moduli & 51 / 49 & 49 / 49 \\
Even/odd counts & 224 / 306 & 306 / 306 \\
Reported answer & $265/357$ & $6/7$ \\
\midrule
Execution & Pass & Pass \\
Program--answer match & Pass & Pass \\
Even-branch test & Pass & Fail \\
Final-answer check & Pass & Fail \\
\bottomrule
\end{tabular}
\par}

\smallskip
\textbf{Counterexample.}
At $n=2$, $\gcd(54,3)=3$, but the faulty reduction
gives $\gcd(3,49)=1$.

\end{minipage}%
}
\caption{Constructed solutions and illustrative checks.}
\label{fig:illustrative_case}
\end{figure}
\section{Conclusion}

The proposed function structured framework connects
mathematical decomposition, executable implementations, and
multi verifier feedback.The framework supports a staged SFT-to-RL training pipeline,
teacher supervision, and structured memory.
The release includes annotated training records and public
evaluation inputs; private verifier specifications remain withheld.

The experiments show that learning valid interfaces is only
a first step toward reliable problem solving. Verifier guided
GRPO improves both final-answer accuracy and complete solution
success. Under the reported inference settings, the trained
Qwen3.5-9B policy exceeds the Qwen3.8-27B API reference across
all reported verification and outcome metrics. Teacher supervision
provides further observed gains, supporting its role as a
complement to verifier guided training.
\clearpage
\begingroup
\interlinepenalty=10000
\bibliographystyle{IEEEbib}
\bibliography{references}
\endgroup
\end{document}